%% file: bare_conf_compsoc.tex
\documentclass[conference]{IEEEtran}
\ifCLASSOPTIONcompsoc
  \usepackage[nocompress]{cite}
\else
  \usepackage{cite}
\fi
\ifCLASSINFOpdf
  \usepackage[pdftex]{graphicx}
\else
  \usepackage[dvips]{graphicx}
\fi
\usepackage{amsmath}
\usepackage{xcolor}
\usepackage{comment}
\usepackage{algorithmic}

\usepackage{booktabs}

\usepackage{array}

\usepackage{subfigure}
\usepackage{fixltx2e}
\usepackage{dblfloatfix}

\usepackage{cleveref}
\usepackage{acronym}
\usepackage{placeins}
\usepackage{parskip}
\usepackage{xspace}
\usepackage{url}

\usepackage[T1]{fontenc}

\usepackage[compact]{titlesec}
\titlespacing{\section}{0pt}{2ex}{1ex}
\titlespacing{\subsection}{0pt}{1ex}{0.5ex}

\titleformat{\subsubsection}[runin]{\normalfont\it}{\thesubsection}{1em}{}

\begin{document}
%
\title{Augmenting data-driven models for energy systems through feature engineering: A Python framework for feature engineering}

\author{Sandra Wilfling }

\input{sources/acronyms}

\maketitle
\setlength{\parskip}{0pt}

\input{sources/abstract.tex}


\input{sources/introduction.tex}
\input{sources/method.tex}

\input{sources/results.tex}
\input{sources/conclusion.tex}

%
\IEEEpeerreviewmaketitle




\bibliographystyle{IEEEtran}
\input{bare_conf_compsoc.bbl}

%



\end{document}

%% file: sources/acronyms.tex
\newacro{CFD}[CFD]{Computational Fluid Dynamics}
\newacro{SVM}[SVM]{Support Vector Machine}
\newacro{HiL}[HiL]{Hardware-in-Loop}
 \newacro{IEEE}[IEEE]{Institute of Electrical and Electronics Engineers}
\newacro{FMI}[FMI]{Functional Mock-up Interface}
\newacro{FMU}[FMU]{Functional Mock-up Unit}
\newacro{HLA}[HLA]{High-Level Architecture}
\newacro{FIR}[FIR]{Finite Impulse Response}
\newacro{IIR}[IIR]{Infinite Impulse Response}
\newacro{CS}[CS]{co-simulation}
\newacro{ML}[ML]{machine learning}
\newacro{WLS}[WLS]{Weighted Least Squares}
\newacro{LSTM}[LSTM]{Long short-term memory}
\newacro{XML}[XML]{Extensible Markup Language}
\newacro{Dymola}[Dymola]{Dassault Systemes Dymola\textsuperscript{\textregistered}}
\newacro{EnergyPlus}[EnergyPlus]{EnergyPlus\texttrademark}
\newacro{MatlabSimulink}[Matlab/Simulink]{MathWorks\textsuperscript{\textregistered} Matlab/Simulink}
\newacro{PCA}[PCA]{Principal Component Analysis}
\newacro{RDC}[RDC]{Randomized Dependence Coefficient}
    \newacro{IEEE}[IEEE]{Institute of Electrical and Electronics Engineers}
    \newacro{FMI}[FMI]{Functional Mock-up Interface}
    \newacro{FMU}[FMU]{Functional Mock-up Unit}
    \newacro{HLA}[HLA]{High-Level Architecture}
    \newacro{CS}[CS]{co-simulation}
    \newacro{ML}[ML]{machine learning}
    \newacro{LSTM}[LSTM]{Long short-term memory}
    \newacro{XML}[XML]{Extensible Markup Language}
    \newacro{Dymola}[Dymola]{Dassault Systemes Dymola\textsuperscript{\textregistered}}
    \newacro{EnergyPlus}[EnergyPlus]{EnergyPlus\texttrademark}
    \newacro{MatlabSimulink}[Matlab/Simulink]{MathWorks\textsuperscript{\textregistered} Matlab/Simulink}
\newacro{AI}[AI]{artificial intelligence}
\newacro{ML}[ML]{machine learning}
\newacro{RF}[RF]{Random Forest}
\newacro{LR}[LR]{Linear Regression}
\newacro{OLS}[OLS]{Ordinary Least Squares}
\newacro{ANN}[ANN]{Artificial Neural Network}
\newacro{FFT}[FFT]{Fast Fourier Transformation}
\newacro{R2}[$R^2$]{Coefficient of Determination}
\newacro{CV-RMS}[$\mathrm{CV\text{-}RMSE}$]{Coefficient of Variation of the Root Mean 
Square Error}
\newacro{RGS}[$\mathrm{RGS}$]{Regression, Gradient guided feature selection} 
\newacro{SCC}[$\mathrm{SCC}$]{Squared Correlation Coefficient} 
\newacro{CV}[$\mathrm{CV}$]{Coefficient of Variation} 
\newacro{MAE}[$\mathrm{MAE}$]{Mean Absolute Error}
\newacro{MAD}[$\mathrm{MAD}$]{Mean Absolute Deviation}

\newacro{MAPE}[$\mathrm{MAPE}$]{Mean Absolute Percentage Error}
\newacro{SVR}[$\mathrm{SVR}$]{Support Vector Regression}

\newacro{kWMD}[$\mathrm{kWMD}$]{kernelized Mahalanobis distance}
\newacro{MSE}[$\mathrm{MSE}$]{mean squared error}
\newacro{RMS}[$\mathrm{RMSE}$]{Root Mean Square Error}
\newacro{VIF}{Variance Inflation Factor}
\newacro{R}[$\mathrm{R}$]{Pearson Correlation Coefficient}
\newacro{CC}[$\mathrm{CC}$-value]{Correlation Coefficient}
\newacro{MIC}[$\mathrm{MIC}$]{Maximum Information Coefficient}
\newacro{OLS}[OLS]{Ordinary Least Squares}
\newacro{HVAC}[HVAC]{Heating, Ventilation, Air Conditioning and Cooling}
\newacro{QQ}[Q-Q]{quantile-quantile}
\newacro{ADF}[ADF]{Augmented Dickey-Fuller}
\newacro{GHG}[GHG]{Greenhouse Gas}
\newacro{EU}[EU]{European Union}
\newacro{CPS}[CPS]{Cyber-Physical System}
\newacro{CPES}[CPES]{Cyber-Physical Energy System}
\newacro{PE}[PE]{Prediction Error}

%% file: sources/abstract.tex
\begin{abstract}
Data-driven modeling is an approach in energy systems modeling that has been gaining popularity. In data-driven modeling, machine learning methods such as linear regression, neural networks or decision-tree based methods are being applied. While these methods do not require domain knowledge, they are sensitive to data quality. Therefore, improving data quality in a dataset is beneficial for creating machine learning-based models.
The improvement of data quality can be implemented through preprocessing methods. A selected type of preprocessing is feature engineering, which focuses on evaluating and improving the quality of certain features inside the dataset. Feature engineering methods include methods such as feature creation, feature expansion, or feature selection. In this work, a Python framework containing different feature engineering methods is presented. This framework contains different methods for feature creation, expansion and selection; in addition, methods for transforming or filtering data are implemented. The implementation of the framework is based on the Python library \textit{scikit-learn}. The framework is demonstrated on a case study of a use case from energy demand prediction. A data-driven model is created including selected feature engineering methods. The results show an improvement in prediction accuracy through the engineered features.

\end{abstract}

%% file: sources/introduction.tex
\section{Introduction}

Modeling and simulation is an crucial step in the design and optimization of energy systems. While traditional modeling methods rely on system parameters, a recent approach focuses on creating data-driven models based on measurement data from an underlying system.
In data-driven modeling, models are not created based on system parameters, but on existing measurement data. These models are based on \ac{ML} methods \cite{Mosavi2019}. While the area of machine learning includes a wide range of methods such as clustering algorithms or classifiers, the focus in data-driven modeling is set to regression analysis for prediction and forecasting \cite{arendtCOMPARATIVEANALYSISWHITE2018}. In regression analysis, methods such as linear regression, decision-tree based regression, or neural networks are being applied \cite{ghofraniPredictionBuildingIndoor2020}. While some of these methods, such as linear regression, can be classified as white-box \ac{ML} methods, others, such as neural networks, are classified as black-box \ac{ML} methods due to their lack of comprehensibility \cite{rudinStopExplainingBlack2019}. While white-box \ac{ML} methods give more insight about their internal structure than black box \ac{ML} methods, their architecture is simpler, making it more difficult to model complex dependencies, for instance non-linearities \cite{loyola2019black}. To capture such dependencies using white-box \ac{ML} models, information about the dependencies can be passed to the model through the dataset. This step is called \textit{feature engineering}.
The main purpose of feature engineering is to augment the existing dataset \cite{kuhnFeatureEngineeringSelection2019}. This can be done through adding new information, or expanding or reducing the existing feature set. In addition, the quality of a single feature can be improved, for instance through transformation or filtering \cite{gomezUseButterworthFilters2001}. 

The area of feature engineering covers a wide number of methods, such as feature expansion \cite{chengPolynomialRegressionAlternative2019} or feature selection \cite{pengFeatureSelectionBased2005}. 
The term \textit{feature creation} covers the creation of features to add new information. Methods of feature creation include encodings of time-based features, such as cyclic features \cite{zhangAccurateForecastingBuilding2020}, or categorical encoding \cite{hancockSurveyCategoricalData2020}.   
Similarly, \textit{feature expansion} is the method of creating new features based on existing features. Feature expansion covers classical methods such as polynomial expansion \cite{chengPolynomialRegressionAlternative2019} or spline interpolation \cite{eilersFlexibleSmoothingBsplines1996}.

In contrast to feature creation and expansion, feature selection aims to reduce the size of the feature set. While large feature sets may contain more information than smaller feature sets, there may be redundancy in the data, as well as sparsity \cite{rothmanSparseMultivariateRegression2010} or multicollinearity \cite{odriscollMitigatingCollinearityLinear2016}. To reduce the sparsity or multicollinearity, as well as to remove redundant features, feature selection mechanisms are applied. 
While methods such as \ac{PCA} \cite{guptaRespiratorySignalAnalysis2016} aim to reduce the feature set through transformation, feature selection methods discard features based on certain criteria \cite{caiFeatureSelectionMachine2018}. Feature selection can be implemented for instance through sequential methods, such as forward or backward selection \cite{guyonIntroductionVariableFeature}, or through correlation criteria \cite{pengFeatureSelectionBased2005}. Correlation criteria include measures based on the \acl{R}, as well as entropy-based criteria \cite{caiFeatureSelectionMachine2018}. The feature selection is then implemented through a threshold-based selection. Threshold-based feature selection analyzes features based on the selected criterion, and discards features below a certain threshold.

Mainly, the methods of feature engineering are applied during the first steps of creating a data-driven model, creating an engineered dataset. This engineered dataset is then used to train the model \cite{chenPhotovoltaicPowerPrediction2021}. However, feature engineering methods can also be used in combination with model selection procedures, such as grid search \cite{akaySupportVectorMachines2009}. 
Feature engineering methods are widely used in applications from the energy domain, such as in prediction for building energy demand \cite{zhengFeatureEngineeringMachine2018 } or photovoltaic power prediction \cite{chenPhotovoltaicPowerPrediction2021}.

\subsection{Main Contribution}
In the creation of data-driven models, a significant factor is the quality of the underlying dataset. To improve the dataset quality, feature engineering methods can be applied. 

The main contribution of this work is a Python framework for feature engineering that can be used for data-driven model creation. The framework implements different methods for feature creation, feature expansion, feature selection or transformation. The feature engineering framework is implemented in Python based on the \textit{scikit-learn} framework and can be imported as a Python package. The functionality of the framework is demonstrated on a case study of an  energy demand prediction use case. The results of the case study show an improvement prediction accuracy through the applied feature engineering steps. 

%% file: sources/method.tex
\section{Method}
The presented framework implements various feature engineering methods in Python based on the research in \cite{polynomialexpansion} and on the interfaces defined by \textit{scikit-learn}. The methods are implemented using either \textit{scikit-learn}'s $TransformerMixin$ or $SelectorMixin$ interface. The framework implements methods for feature expansion, feature creation, feature selection, as well as transformation and filtering operations.

\subsection{Feature Creation and Expansion}
In the framework, different methods for feature creation and expansion are implemented. These methods create new features from time values or from expansion of existing features.
To create new features, the implemented framework supports categorical encoding and cyclic encoding of time-based values.
\subsubsection*{Cyclic Features} 
Cyclic features can be used to model time values through cyclic functions \cite{zhangAccurateForecastingBuilding2020}. Cyclic features were implemented in \cite{polynomialexpansion}, as well as in  \cite{doglianiMachineLearningBuilding} and  \cite{schranzMachineLearningWater2020}.  
In the implementation of the framework, sinusoidal signals $x_{sin}, x_{cos}$ with a selected frequency $f$ can be created based on a sample series $n$:
\begin{eqnarray}
    x_{sin}[n] = sin(2 \pi f n)\\
    x_{cos}[n] = cos(2 \pi f n)
\end{eqnarray}
The implementation offers the creation of features with a zero-order hold function for a certain time period, for instance $T_S = 1~day$ for a signal with a time period of $T=1~week$. 
\subsubsection*{Categorical Features}
Categorical encoding creates a representation of discrete numerical values through a number of features with boolean values \cite{hancockSurveyCategoricalData2020,polynomialexpansion}.
In this implementation, for a number of categorical features $x_{0,....,N}$ for a feature $x$ with discrete possible values $v_{0,....,N}$, a single feature $x_{i}$ is defined as:
\begin{equation}
    x_{i} = \begin{cases}
            1 & x = v_{i}\\
            0 & else
            \end{cases}
\end{equation}
The framework offers categorical encoding for time-based values.
In addition, a division factor is implemented to create an encoding of a downsampled version of the time values. 
\subsubsection*{Feature Expansion}
For feature expansion, the framework implements wrappers for \textit{scikit-learn}'s \textit{PolynomialFeatures} and \textit{SplineTransformer} classes. The method of polynomial expansion was applied in \cite{polynomialexpansion}.  The parameters for the expansion methods are passed through the wrapper.
\subsubsection*{Time-based Features}
The framework implements a method of dynamic timeseries unrolling to create features $x_{n-1}$, $x_{n-2}$, ... $x_{n-N}$ from an existing feature \textit{x}. The method of dynamic timeseries unrolling is based on the research in \cite{falayCouplingPhysicalMachine2021, schranzEnergyPredictionChanged2021}, and \cite{doglianiMachineLearningBuilding}. While \cite{falayCouplingPhysicalMachine2021} and \cite{schranzEnergyPredictionChanged2021} use dynamic timeseries unrolling for both input and target features of a model, 
allowing the creation of auto-recursive models, this implementation only supports dynamic timeseries unrolling for the input features, similar to the method used in \cite{doglianiMachineLearningBuilding}. In this implementation, dynamic timeseries unrolling is implemented through filter operations from the \textit{scipy.signal} library. The dynamic features are created through the convolution of the signal $x$ with a Kronecker delta for $i = 1...N$:
\begin{equation}
    x_{dyn,i}[n] = x[n] * \delta[n-i] 
\end{equation}
This operation creates delayed signals $x_{dyn,1},...,x_{dyn,N}$. In our implementation, for the samples in the delayed signals, for which no values are available, zero values are used. 
\subsection{Feature Selection}
In the framework, several threshold-based feature selection methods are implemented. These methods analyze the input and target features based on a certain criterion, and then discard features with a low value of the criterion. 
A widely used criterion is the \acl{R}, which is used to detect linear correlations between features \cite{chenPhotovoltaicPowerPrediction2021}. The \acl{R} calculates the correlation between two features for samples $x_{0,....,N}, y_{0,...,N}$ with mean values $\bar{x}$ and $\bar{y}$: 
\begin{equation}
    r_{x,y} = \frac{\sum_{i=0}^{N}(x_i - \bar{x})(y_i - \bar{y})}
    {\sqrt{\sum_{i=0}^N (x_i - \bar{x})^2\sum_{i=0}^N(y_i - \bar{y})^2}}
\end{equation}
While the Pearson correlation identifies linear correlations, non-linear dependencies are not detected. To detect non-linear dependencies, criteria such as \ac{MIC}\cite{reshefMeasuringDependencePowerfully}, \textit{ennemi}\cite{laarneEnnemiNonlinearCorrelation2021}, \textit{dCor} \cite{szekelyMeasuringTestingDependence2007} or the \ac{RDC}\cite{lopez-pazRandomizedDependenceCoefficient} can be used.

The framework provides classes for the criteria \acl{R}, F-statistic based on the \acl{R}, as well as thresholds based on the \ac{MIC}, \textit{ennemi} and \ac{RDC}.  

\subsection{Transformation and Filtering Operations}
To transform features, the framework implements the \textit{Box-cox} transformation as well as the square root and inverse transformation. 
In addition, the framework provides filtering operations, which were applied in timeseries prediction for instance in \cite{gomezUseButterworthFilters2001}.
Discrete-time based filters can be implemented in Python through the functions implemented in \textit{scipy.signal}. The \textit{scipy.signal} library offers functions for calculating the coefficients for different types of digital filters. A digital filter of order $N$ can be defined through the transfer function $H(z)$ in a direct form:
\begin{equation}
    H(z) = \frac{\sum_{i=0}^{N} b_i z^{i}}{\sum_{i=0}^{N} a_i z^{i}}
\end{equation}
The filter coefficients $a_i$ and $b_i$ define the behavior of the filter.
The \textit{scipy.signal} library offers functions to compute the filter coefficients for filter types such as the Butterworth or Chebyshev filter \cite{sandhuStudyDesignImplementation2016}. While \textit{scipy.signal} offers the computation of analog and digital filter coefficients, the framework implementation focuses on digital filter implementations. The framework implements the Butterworth and Chebyshev filter as \textit{scikit-learn} \textit{TransformerMixin} classes. 
In addition, an envelope detection filter was implemented for demodulation of modulated signals. This filter was implemented using the \textit{pandas} rolling average function. For all filters, offset compensation before and after applying the filter operation and a mask for handling \textit{NaN} values were implemented. 
The direct form filter classes of the framework offer a simple option for extension. Different architectures can be implemented by re-defining the implemented method for coefficient calculation. This allows to create filters with different \ac{FIR} or \ac{IIR} structures.


\subsection{Composite Transformers}
In feature engineering, it is often the case that only a selected subset of features should be transformed. To offer the possibility to transform only selected features, a composite transformer wrapper was implemented. This wrapper offers to either automatically replace features through their transformed versions, or add transformed features separately to the dataset. 
\subsection{Implementation}
The framework offers compatibility with the \textit{sklearn.Pipeline} implementation, making it possible to use objects as part of a \ac{ML} pipeline. The parameters of each objects can be adapted through grid search, for instance using \textit{sklearn.model\_selection.GridSearchCV}. In addition, every created object can be stored to and loaded from a Pickle file using the \textit{save\_pkl} or \textit{load\_pkl} method.

While the filtering, feature expansion and feature creation methods support operations on a \textit{numpy.ndarray} or \textit{pd.Dataframe} or \textit{pd.Series} object, the feature creation methods require a \textit{pd.Dataframe} or \textit{pd.Series} object with a \textit{DateTimeIndex} or \textit{TimedeltaIndex} to create samples based on a certain date.

%% file: sources/results.tex
\section{Case Study}
The framework is demonstrated on a use case from prediction for energy systems modeling. For this purpose, a mixed office-campus building is selected. A prediction model should be trained based on existing measurement data. The data-driven model is created using a workflow based on the implemented methods. 

\subsection{Application}
In this case study, the energy demand of a mixed office-campus building should be evaluated. The data was provided from the research in \cite{schranzEnergyPredictionChanged2021}.
The energy demand of a building is subject to various factors.
Main factors that influence building energy demand are thermal characteristics and \ac{HVAC} system behavior \cite{maccariniDevelopmentModelicabasedSimplified2021}. Additionally, building energy demand may be dependent on occupancy \cite{ghofraniPredictionBuildingIndoor2020} or subject to seasonal trends \cite{zhangAccurateForecastingBuilding2020}. Many of these factors show non-linear behavior, which makes it difficult to address them through a purely linear model. Therefore, feature engineering was used to model additional factors.

\subsection{Data-driven Model}
For the selected application, a data-driven model of the building energy demand should be created. To demonstrate the effect of feature engineering, two models were trained based on the existing measurement data: a basic regression model and a regression model with engineered features.
\subsubsection*{Measurement Data}
The energy demand was measured during a period from 05/2019 to 03/2020, with a sampling time of $1h$ \cite{schranzEnergyPredictionChanged2021}. The measurement data includes features based on weather data, such as \textit{temperature}, as well as occupancy data, such as \textit{registrations}. The rest of the features are time-based, such as \textit{daytime} or \textit{weekday}.
\begin{table}[h]
\centering
\caption{Feature Set for Energy Consumption Prediction}
  \begin{tabular}{  l c l}
        \toprule
        \textbf{Feature Name} & \textbf{Unit} & \textbf{Description}\\
        \midrule
        \textit{temperature} & \textdegree C & Outdoor Temperature \\
        \textit{daytime} & h & Daytime \\
        \textit{weekday} & d & Weekday from 0 to 6\\
        \textit{holiday} & & Public holiday\\
        \textit{daylight} &  & day or night \\
        \textit{registrations} & & registrations for lectures\\
        \midrule
        Consumption & kWh & Energy Consumption\\
        \bottomrule
    \end{tabular}
\end{table}

\subsubsection*{Model Architecture}
For the energy demand, a linear regression model should be trained. The linear regression architecture was selected due to its simplicity and comprehensibility as a white-box \ac{ML} model. Non-linear behavior of the underlying system should be incorporated through feature engineering.
\subsubsection*{Feature Engineering}
\begin{figure}[h]
    \centering
    \vspace{-8mm}
    \includegraphics[width=\columnwidth]{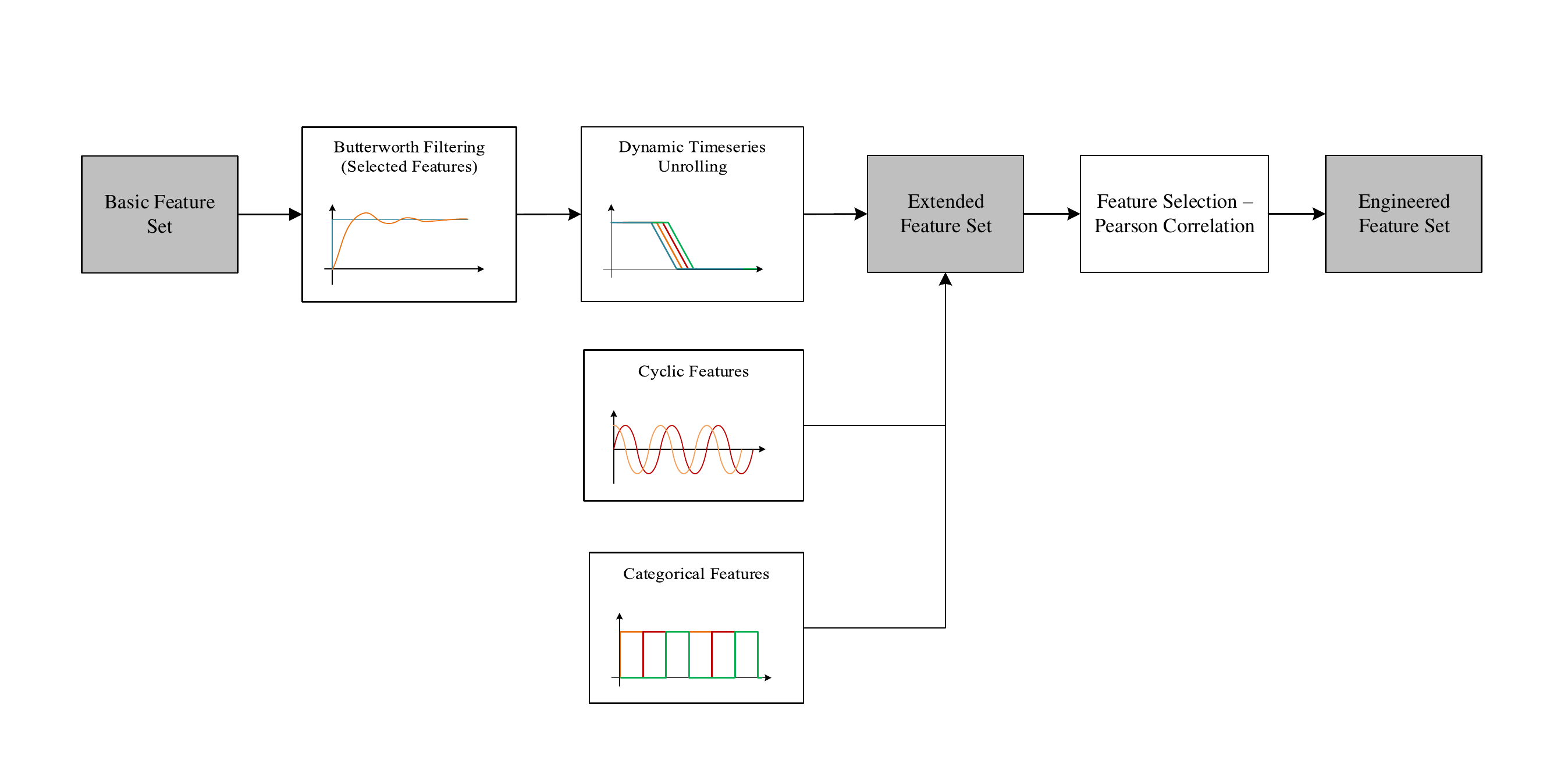}
    \vspace{-8mm}
    \caption{Implemented Workflow.}
    \label{fig:method:workflow}
    \vspace{-3mm}
\end{figure}

To model the non-linear behavior of the energy demand, categorical features and cyclical features were used in combination with Butterworth Filtering, dynamic timeseries unrolling and feature selection through the \acl{R}. An overview of the implemented workflow is depicted in Figure \ref{fig:method:workflow}.
\subsubsection*{Training Parameters}
For the model training, a train-test split of 0.8 was selected together with a 5-fold cross-validation. For the model with engineered features, the parameters for the steps timeseries unrolling and feature selection were determined through a grid search based on the metrics \ac{R2}, \ac{MSE} and \ac{MAPE}.

\subsection{Experimental Results}
 The two models were trained on the measurement data and compared in terms of performance metrics. Additionally, analyses of the predicted values through timeseries analysis and prediction error plots were performed.
\subsubsection*{Performance Metrics}
To evaluate the performance of the model, the metrics \ac{R2}, \ac{CV-RMS} and \ac{MAPE} were used \cite{polynomialexpansion}. Table \ref{tab:res:metr} gives an overview of the metrics.

\begin{table}[h]
    \centering
    \caption{Performance Metrics}
    \begin{tabular}{l | c c c }
    \toprule
     Model &  \ac{R2} & \ac{CV-RMS} & \ac{MAPE}  \\
     \midrule
     Basic Regression & 0.548 & 0.267 & 22.764 \%
\\
     Engineered Features & 0.638 & 0.201 & 17.493\%\\
     \bottomrule
    \end{tabular}
    \label{tab:res:metr}
\end{table}
From the performance metrics, an improvement in prediction accuracy for the linear regression model through the engineered features could be observed. 

\subsubsection*{Timeseries Analysis}
\begin{figure}[!h]
    \centering
    \includegraphics[width=\columnwidth,page=1]{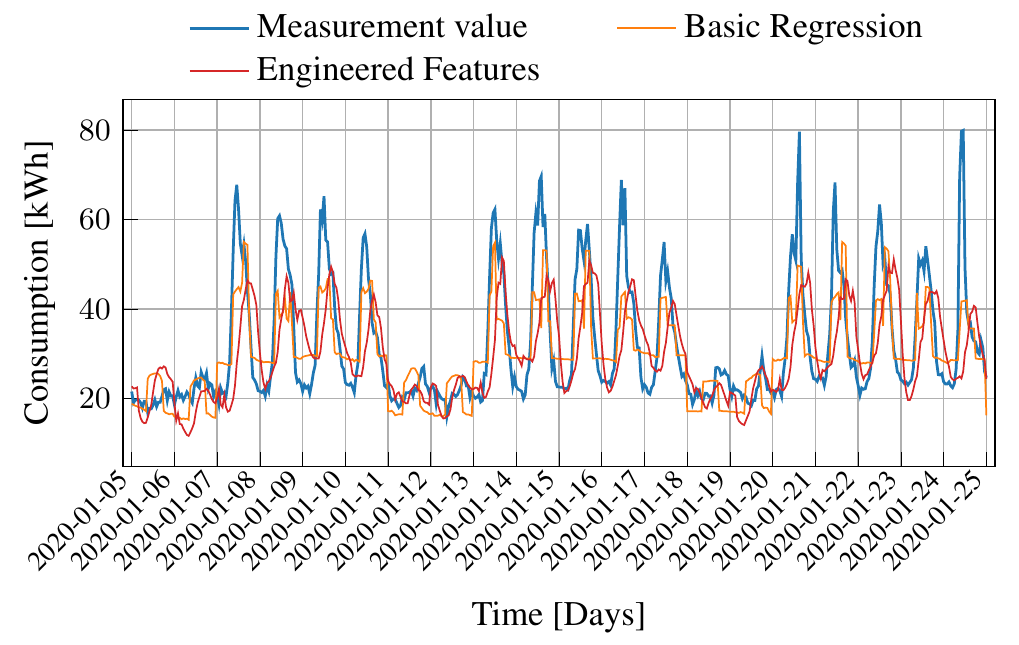}
    \vspace{-8mm}
    \caption{Timeseries Analysis for period of 25 days from test set.}
    \label{fig:res:timeseries}
  
\end{figure}

The improvement in prediction accuracy could also be observed from the timeseries analysis depicted in Figure \ref{fig:res:timeseries}. 

The timeseries analysis showed that the cyclic behavior of the day-night changes in the energy demand could be more accurately replicated by the model with engineered features.

\begin{figure}[!h]
    \centering
    \includegraphics[width=\columnwidth,page=2]{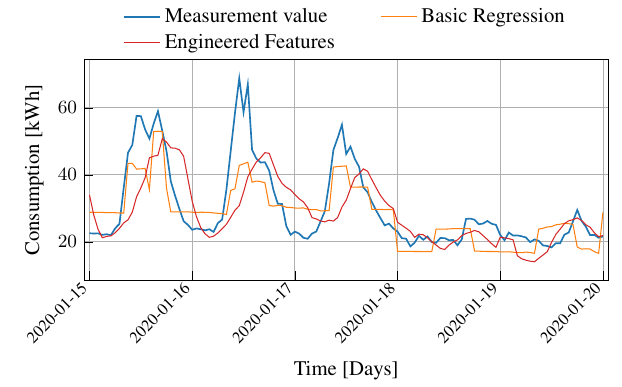}
    \vspace{-8mm}
    \caption{Timeseries Analysis for period of five days from test set.}
    \label{fig:res:timeseries:close}
\end{figure}

Additionally, the prediction using engineered features shows a higher accuracy in replicating low energy demand values than the basic regression. This effect can be observed in Figure \ref{fig:res:timeseries:close}.

\begin{figure}[h]
\centering
\begin{subfigure}
\centering
    \includegraphics[width=0.45\columnwidth, page=1]{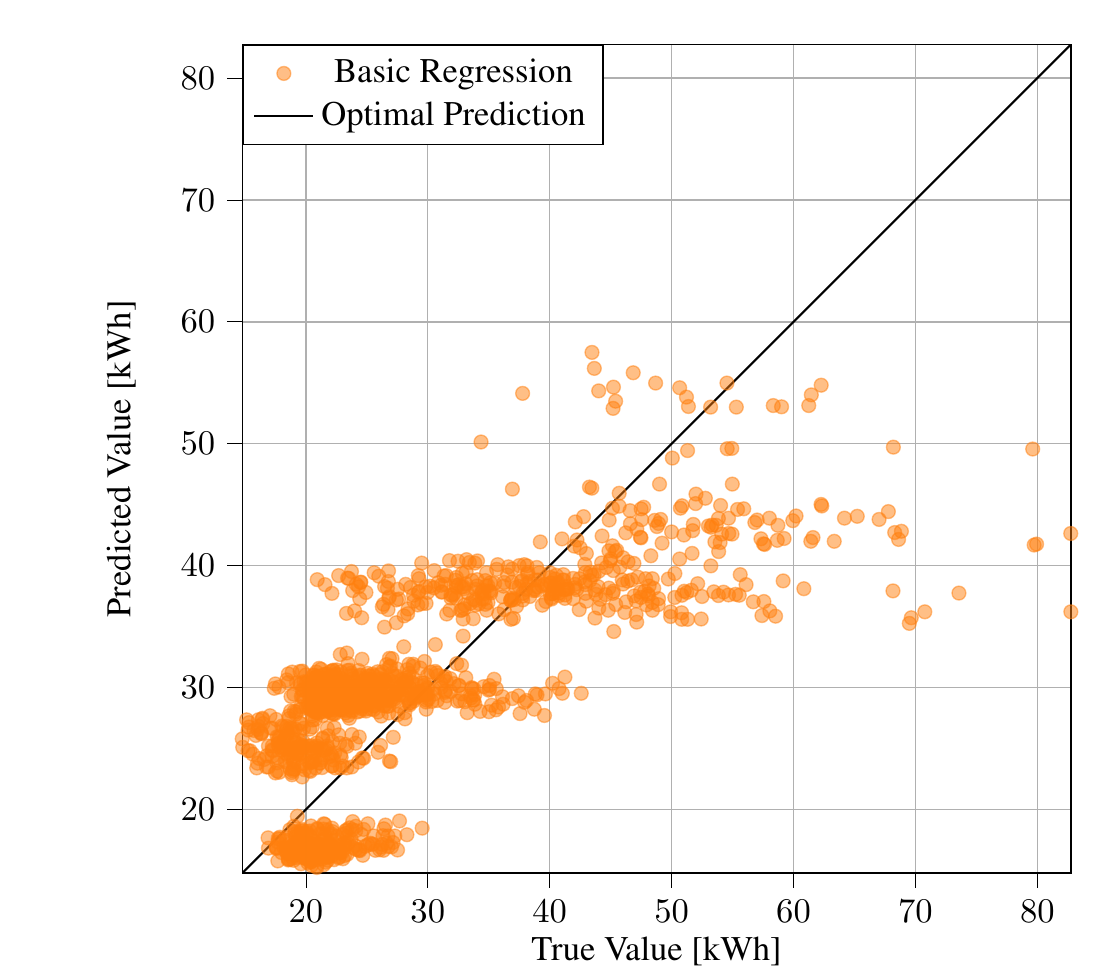}
    \end{subfigure}
\begin{subfigure}
\centering
    \includegraphics[width=0.45\columnwidth,page=2]{Figures/Figures_PDF/scatter.pdf}
\end{subfigure}
\caption{Prediction Error Plots for Energy Consumption}
\vspace{-3mm}
\label{fig:res:scatter}
\end{figure}

For both models, the residual error was analyzed through prediction error plots (Figure \ref{fig:res:scatter}). The prediction error plots show that the residual error is decreased for the model with engineered features. In addition, the homogenity of the error distribution is improved through the applied feature engineering methods.

Since performance metrics, timeseries analysis and prediction error plots show an improvement in accuracy, the feature engineering steps are suggested to be beneficial for the prediction model.

%% file: sources/conclusion.tex
\section{Related Work}

In the creation of data-driven models in Python, many frameworks have been implemented. One of the most well-known Python \ac{ML} frameworks is the \textit{scikit-learn} framework, which provides methods such as data preprocessing, feature engineering, clustering, and implementations of various \ac{ML} models. The \textit{scikit-learn} framework offers interfaces which can be used to implement additional methods.
Due to the popularity of \textit{scikit-learn}, various frameworks extending \textit{scikit-learn} have been implemented. 
For instance, the \textit{imblearn} framework \cite{lemaitre2017imblearn} focuses on extending \textit{scikit-learn}'s functionality to processing imbalanced datasets. In addition, the \textit{imblearn} framework offers different resampling methods.
The \textit{mlxtend} framework \cite{raschkas_2018_mlxtend} offers feature extraction methods such as \ac{PCA}, or feature selection methods such as sequential feature selection. Additionally, different evaluation and utility functions are implemented.
In contrast, libraries such as \textit{statsmodels}\cite{seabold2010statsmodels} provide their own interface for their regression models. The \textit{statsmodels} framework provides models based on stochastic and statistical methods, such as the \ac{WLS}. 
In the area of feature engineering, different Python packages have been created. The \textit{feature-engine} \cite{galliFeatureenginePythonPackage2021} library contains a large collection of feature engineering methods, which are implemented based on \textit{scikit-learn}. The \textit{featuretools} framework \cite{kanter2015deep} allows the synthesis of features from relational databases. offers functionality for feature encoding, as well as different transformations or aggregate functions. Additionally, this framework offers transformations, feature encoding, aggregate functions, as well as coordinate transformations.
\section{Conclusion} 
This paper presents a Python framework for feature engineering that provides different methods through a standardized interface. The framework is based on the \textit{scikit-learn} package and offers different methods. The framework offers  classic feature engineering methods such feature expansion, as well as as feature creation, feature selection or transformation and filter operations.
The framework is implemented as a Python package and can be included in different projects. Through the specifically defined interfaces of the framework, additional methods can be added with low effort. Finally, we demonstrate the framework on a case study of energy demand prediction, using a workflow created from a subset of the implemented methods for data-driven model creation.
\subsection{Future Work}
The current version of the framework gives many options for extensions. For instance, additional feature engineering methods can be added using the provided interfaces of the framework. In addition, combinations of the implemented feature engineering methods can be used for prediction in different use cases.

%% file: bare_conf_compsoc.bbl

%% file: Featureengineering/bib/bib_full.bib
@article{Mosavi2019,
  doi = {10.3390/en12071301},
  url = {https://doi.org/10.3390/en12071301},
  year = {2019},
  month = apr,
  publisher = {{MDPI} {AG}},
  volume = {12},
  number = {7},
  pages = {1301},
  author = {Amir Mosavi and Mohsen Salimi and Sina Faizollahzadeh Ardabili and Timon Rabczuk and Shahaboddin Shamshirband and Annamaria Varkonyi-Koczy},
  title = {State of the Art of Machine Learning Models in Energy Systems,  a Systematic Review},
  journal = {Energies}
}

@inproceedings{arendtCOMPARATIVEANALYSISWHITE2018,
  title = {{{Comparative Analysis of white-}}, {{gray- and black-box models for thermal simulation of indoor environment}}: {{Teaching Building Case Study}}},
  booktitle = {2018 {{Building Performance Modeling Conference}} and {{SimBuild}} Co-Organized by {{ASHRAE}} and {{IBPSA-USA Chicago}}},
  author = {Arendt, K and Jradi, M and Shaker, H R and Veje, C T},
  year = {2018},
  pages = {8},
  langid = {english}
}

@article{ghofraniPredictionBuildingIndoor2020,
  title = {Prediction of Building Indoor Temperature Response in Variable Air Volume Systems},
  author = {Ghofrani, Ali and Nazemi, Seyyed Danial and Jafari, Mohsen A.},
  year = {2020},
  month = jan,
  journal = {Journal of Building Performance Simulation},
  volume = {13},
  number = {1},
  pages = {34--47},
  issn = {1940-1493, 1940-1507},
  doi = {10.1080/19401493.2019.1688393},
  language = {en},
}

@article{rudinStopExplainingBlack2019,
  title = {Stop Explaining Black Box Machine Learning Models for High Stakes Decisions and Use Interpretable Models Instead},
  author = {Rudin, Cynthia},
  year = {2019},
  month = may,
  journal = {Nature Machine Intelligence},
  volume = {1},
  number = {5},
  pages = {206--215},
  issn = {2522-5839},
  doi = {10.1038/s42256-019-0048-x},
  langid = {english}
}

@article{loyola2019black,
  title = {Black-Box vs. White-Box: {{Understanding}} Their Advantages and Weaknesses from a Practical Point of View},
  author = {{Loyola-Gonzalez}, Octavio},
  year = {2019},
  journal = {IEEE access : practical innovations, open solutions},
  volume = {7},
  pages = {154096--154113},
  publisher = {{IEEE}}
}

@book{kuhnFeatureEngineeringSelection2019,
  title = {Feature {{Engineering}} and {{Selection}}: {{A Practical Approach}} for {{Predictive Models}}},
  shorttitle = {Feature {{Engineering}} and {{Selection}}},
  author = {Kuhn, Max and Johnson, Kjell},
  year = {2019},
  month = jul,
  publisher = {{CRC Press}},
  googlebooks = {q5alDwAAQBAJ},
  isbn = {978-1-351-60946-3},
  langid = {english}
}

@article{gomezUseButterworthFilters2001,
  title = {The {{Use}} of {{Butterworth Filters}} for {{Trend}} and {{Cycle Estimation}} in {{Economic Time Series}}},
  author = {G{\'o}mez, V{\'i}ctor},
  year = {2001},
  month = jul,
  journal = {Journal of Business \& Economic Statistics},
  volume = {19},
  number = {3},
  pages = {365--373},
  publisher = {{Taylor \& Francis}},
  issn = {0735-0015},
  doi = {10.1198/073500101681019909}
}

@article{chengPolynomialRegressionAlternative2019,
  title = {Polynomial {{Regression As}} an {{Alternative}} to {{Neural Nets}}},
  author = {Cheng, Xi and Khomtchouk, Bohdan and Matloff, Norman and Mohanty, Pete},
  year = {2019},
  month = apr,
  journal = {arXiv:1806.06850 [cs, stat]},
  eprint = {1806.06850},
  eprinttype = {arxiv},
  primaryclass = {cs, stat},
  archiveprefix = {arXiv},
  langid = {english}
}

@article{pengFeatureSelectionBased2005,
  title = {Feature Selection Based on Mutual Information: {{Criteria}} of {{Max-Dependency}}, {{Max-Relevance}}, and {{Min-Redundancy}}},
  shorttitle = {Feature Selection Based on Mutual Information},
  author = {Peng, H. and Long, F. and Ding, C.},
  year = {2005},
  journal = {IEEE Transactions on Pattern Analysis and Machine Intelligence},
  volume = {27},
  number = {8},
  pages = {1226--1238},
  issn = {0162-8828},
  doi = {10.1109/TPAMI.2005.159},
  langid = {english}
}

@article{zhangAccurateForecastingBuilding2020,
  title = {Accurate Forecasting of Building Energy Consumption via a Novel Ensembled Deep Learning Method Considering the Cyclic Feature},
  author = {Zhang, Guiqing and Tian, Chenlu and Li, Chengdong and Zhang, Jun Jason and Zuo, Wangda},
  year = {2020},
  month = jun,
  journal = {Energy},
  volume = {201},
  pages = {117531},
  issn = {0360-5442},
  doi = {10.1016/j.energy.2020.117531},
  langid = {english}
}

@article{hancockSurveyCategoricalData2020,
  title = {Survey on Categorical Data for Neural Networks},
  author = {Hancock, John T. and Khoshgoftaar, Taghi M.},
  year = {2020},
  month = dec,
  journal = {Journal of Big Data},
  volume = {7},
  number = {1},
  pages = {28},
  issn = {2196-1115},
  doi = {10.1186/s40537-020-00305-w},
  langid = {english}
}

@article{eilersFlexibleSmoothingBsplines1996,
  title = {Flexible Smoothing with {{B-splines}} and Penalties},
  author = {Eilers, Paul H. C. and Marx, Brian D.},
  year = {1996},
  month = may,
  journal = {Statistical Science},
  volume = {11},
  number = {2},
  issn = {0883-4237},
  doi = {10.1214/ss/1038425655}, 
  langid = {english},
}

@article{rothmanSparseMultivariateRegression2010,
  title = {Sparse {{Multivariate Regression With Covariance Estimation}}},
  author = {Rothman, Adam J. and Levina, Elizaveta and Zhu, Ji},
  year = {2010},
  month = jan,
  journal = {Journal of Computational and Graphical Statistics},
  volume = {19},
  number = {4},
  pages = {947--962},
  issn = {1061-8600, 1537-2715},
  doi = {10.1198/jcgs.2010.09188},
  langid = {english}
}

@article{odriscollMitigatingCollinearityLinear2016,
  title = {Mitigating Collinearity in Linear Regression Models Using Ridge, Surrogate and Raised Estimators},
  author = {O'Driscoll, Diarmuid and Ramirez, Donald},
  year = {2016},
  month = jan,
  journal = {Cogent Mathematics},
  volume = {3},
  pages = {1144697},
  doi = {10.1080/23311835.2016.1144697}
}

@inproceedings{guptaRespiratorySignalAnalysis2016,
  title = {Respiratory Signal Analysis Using {{PCA}}, {{FFT}} and {{ARTFA}}},
  booktitle = {2016 {{International Conference}} on {{Electrical Power}} and {{Energy Systems}} ({{ICEPES}})},
  author = {Gupta, Varun and Mittal, Monika},
  year = {2016},
  month = dec,
  pages = {221--225},
  doi = {10.1109/ICEPES.2016.7915934},
  langid = {english},
}

@article{caiFeatureSelectionMachine2018,
  title = {Feature Selection in Machine Learning: {{A}} New Perspective},
  shorttitle = {Feature Selection in Machine Learning},
  author = {Cai, Jie and Luo, Jiawei and Wang, Shulin and Yang, Sheng},
  year = {2018},
  month = jul,
  journal = {Neurocomputing},
  volume = {300},
  pages = {70--79},
  issn = {09252312},
  doi = {10.1016/j.neucom.2017.11.077},
  langid = {english}
}

@article{guyonIntroductionVariableFeature,
  title = {An {{Introduction}} to {{Variable}} and {{Feature Selection}}},
    author={Guyon, Isabelle and Elisseeff, Andr{\'e}},
  journal={Journal of machine learning research},
  volume={3},
  number={Mar},
  pages={1157--1182},
  year={2003}
}

@article{chenPhotovoltaicPowerPrediction2021,
  title = {Photovoltaic Power Prediction of {{LSTM}} Model Based on {{Pearson}} Feature Selection},
  author = {Chen, Hailang and Chang, Xianfa},
  year = {2021},
  month = nov,
  journal = {Energy Reports},
  series = {2021 {{International Conference}} on {{Energy Engineering}} and {{Power Systems}}},
  volume = {7},
  pages = {1047--1054},
  issn = {2352-4847},
  doi = {10.1016/j.egyr.2021.09.167},
  langid = {english}
}

@article{akaySupportVectorMachines2009,
  title = {Support Vector Machines Combined with Feature Selection for Breast Cancer Diagnosis},
  author = {Akay, Mehmet Fatih},
  year = {2009},
  month = mar,
  journal = {Expert Systems with Applications},
  volume = {36},
  number = {2},
  pages = {3240--3247},
  issn = {09574174},
  doi = {10.1016/j.eswa.2008.01.009},
  langid = {english},
}

@book{zhengFeatureEngineeringMachine2018,
  title = {Feature {{Engineering}} for {{Machine Learning}}: {{Principles}} and {{Techniques}} for {{Data Scientists}}},
  shorttitle = {Feature {{Engineering}} for {{Machine Learning}}},
  author = {Zheng, Alice and Casari, Amanda},
  year = {2018},
  edition = {1st},
  publisher = {{O'Reilly Media, Inc.}},
  isbn = {978-1-4919-5324-2}
}

@inproceedings{polynomialexpansion,
  title = {Learning Non-linear White-box Predictors: A Use Case in Energy Systems},
  author = {Wilfling, Sandra and Ebrahimi, Masoud and Alfalouji, Qamar and Schweiger, Gerald and Basirat, Mina},
  year = {2022},
  booktitle={21st IEEE International Conference on Machine Learning and Applications},
  publisher = {{IEEE}}
}

@article{doglianiMachineLearningBuilding,
  title = {Machine {{Learning}} for {{Building Energy Prediction}}: {{A Case Study}} of an {{Office Building}}},
  author = {Dogliani, Matias and Nord, Nathan and Doblas, {\'A}ngeles and Calixto, Ian and Wilfling, Sandra and Alfalouji, Qamar and Schweiger, Gerald},
  pages = {8},
  langid = {english}
}

@incollection{schranzMachineLearningWater2020,
  title = {Machine {{Learning}} for {{Water Supply Supervision}}},
  booktitle = {Trends in {{Artificial Intelligence Theory}} and {{Applications}}. {{Artificial Intelligence Practices}}},
  author = {Schranz, Thomas and Schweiger, Gerald and Pabst, Siegfried and Wotawa, Franz},
  editor = {Fujita, Hamido and {Fournier-Viger}, Philippe and Ali, Moonis and Sasaki, Jun},
  year = {2020},
  volume = {12144},
  pages = {238--249},
  publisher = {{Springer International Publishing}},
  address = {{Cham}},
  doi = {10.1007/978-3-030-55789-8_21},
  isbn = {978-3-030-55788-1 978-3-030-55789-8},
  langid = {english}
}

@article{falayCouplingPhysicalMachine2021,
  title = {Coupling Physical and Machine Learning Models: Case Study of a Single-Family House},
  shorttitle = {Coupling Physical and Machine Learning Models},
  author = {Falay, Basak and Wilfling, Sandra and Alfalouji, Qamar and Exenberger, Johannes and Schranz, Thomas and Legaard, Christian M{\o}ldrup and Leusbrock, Ingo and Schweiger, Gerald},
  year = {2021},
  month = sep,
  journal = {Modelica Conferences},
  pages = {335--341},
  issn = {1650-3740},
  doi = {10.3384/ecp21181335},
  copyright = {Copyright (c) 2021 Array},
  langid = {american},
}

@article{reshefMeasuringDependencePowerfully,
  title = {Measuring {{Dependence Powerfully}} and {{Equitably}}},
  author = {Reshef, Yakir A and Reshef, David N and Finucane, Hilary K and Sabeti, Pardis C and Mitzenmacher, Michael},
  pages = {63},
  year={2016},
  journal={Journal of Machine Learning Research},
  langid = {english},
}

@article{laarneEnnemiNonlinearCorrelation2021,
  title = {Ennemi: {{Non-linear}} Correlation Detection with Mutual Information},
  shorttitle = {Ennemi},
  author = {Laarne, Petri and Zaidan, Martha A. and Nieminen, Tuomo},
  year = {2021},
  month = jun,
  journal = {SoftwareX},
  volume = {14},
  pages = {100686},
  issn = {2352-7110},
  doi = {10.1016/j.softx.2021.100686},
  langid = {english}
}

@article{szekelyMeasuringTestingDependence2007,
  title = {Measuring and Testing Dependence by Correlation of Distances},
  author = {Sz{\'e}kely, G{\'a}bor J. and Rizzo, Maria L. and Bakirov, Nail K.},
  year = {2007},
  month = dec,
  journal = {The Annals of Statistics},
  volume = {35},
  number = {6},
  eprint = {0803.4101},
  eprinttype = {arxiv},
  primaryclass = {math, stat},
  issn = {0090-5364},
  doi = {10.1214/009053607000000505},
  archiveprefix = {arXiv},
  langid = {english}
}

@article{lopez-pazRandomizedDependenceCoefficient,
  title = {The {{Randomized Dependence Coefficient}}},
  author = {{Lopez-Paz}, David and Hennig, Philipp and Sch{\"o}lkopf, Bernhard},
  pages = {9},
  langid = {english}
}

@article{sandhuStudyDesignImplementation2016,
  title = {A {{Study}} on {{Design}} and {{Implementation}} of {{Butterworth}}, {{Chebyshev}} and {{Elliptic Filter}} with {{MatLab}}},
  author = {Sandhu, Manjit and Kaur, Sukhdeep and Kaur, Jaipreet},
  year = {2016},
  volume = {4},
  number = {6},
  pages = {4},
  langid = {english}
}

@article{maccariniDevelopmentModelicabasedSimplified2021,
  title = {Development of a {{Modelica-based}} Simplified Building Model for District Energy Simulations},
  author = {Maccarini, Alessandro and Prataviera, Enrico and Zarrella, Angelo and Afshari, Alireza},
  year = {2021},
  month = nov,
  journal = {Journal of Physics: Conference Series},
  volume = {2042},
  number = {1},
  pages = {012078},
  issn = {1742-6588, 1742-6596},
  doi = {10.1088/1742-6596/2042/1/012078},
  langid = {english},
}

@article{lemaitre2017imblearn,
author  = {Guillaume  Lema{{\^i}}tre and Fernando Nogueira and Christos K. Aridas},
title   = {Imbalanced-learn: A Python Toolbox to Tackle the Curse of Imbalanced Datasets in Machine Learning},
journal = {Journal of Machine Learning Research},
year    = {2017},
volume  = {18},
number  = {17},
pages   = {1-5},
url     = {http://jmlr.org/papers/v18/16-365}
}

@article{raschkas_2018_mlxtend,
  author       = {Sebastian Raschka},
  title        = {MLxtend: Providing machine learning and data science 
                  utilities and extensions to Python’s  
                  scientific computing stack},
  journal      = {The Journal of Open Source Software},
  volume       = {3},
  number       = {24},
  month        = apr,
  year         = 2018,
  publisher    = {The Open Journal},
  doi          = {10.21105/joss.00638},
  url          = {http://joss.theoj.org/papers/10.21105/joss.00638}
}

@inproceedings{seabold2010statsmodels,
  title={statsmodels: Econometric and statistical modeling with python},
  author={Seabold, Skipper and Perktold, Josef},
  booktitle={9th Python in Science Conference},
  year={2010},
}

@article{galliFeatureenginePythonPackage2021,
  title = {Feature-Engine: {{A Python}} Package for Feature Engineering for Machine Learning},
  shorttitle = {Feature-Engine},
  author = {Galli, Soledad},
  year = {2021},
  month = sep,
  journal = {Journal of Open Source Software},
  volume = {6},
  number = {65},
  pages = {3642},
  issn = {2475-9066},
  doi = {10.21105/joss.03642},
  langid = {english}
}

@inproceedings{kanter2015deep,
  author    = {James Max Kanter and Kalyan Veeramachaneni},
  title     = {Deep feature synthesis: Towards automating data science endeavors},
  booktitle = {2015 {IEEE} International Conference on Data Science and Advanced Analytics, DSAA 2015, Paris, France, October 19-21, 2015},
  pages     = {1--10},
  year      = {2015},
  organization={IEEE}
}
